\documentclass[5p,times,procedia]{elsarticle}
\usepackage{ecrc}

\volume{00}

\firstpage{1}

\journalname{Procedia CIRP}

\runauth{B. Alt et al.}

\jid{trpro}

\usepackage{amssymb}

\usepackage[figuresright]{rotating}

\usepackage[nolist]{acronym}
\usepackage{comment}
\usepackage{todonotes}
\usepackage[inkscapearea=drawing, inkscapepath=figures/generated, inkscapelatex=false]{svg}
\usepackage{amsmath}

\acrodef{xui}[XUI]{Explanation User Interface}
\acrodef{ai}[AI]{artificial intelligence}
\acrodef{xai}[XAI]{explainable artificial intelligence}
\acrodef{ERP}[ERP]{Enterprise Resource Planning}
\acrodef{MES}[MES]{Manufacturing Execution System}
\acrodef{dl}[DL]{deep learning}
\acrodef{ui}[UI]{user interface}
\acrodef{spi}[SPI]{Shadow Program Inversion}
\acrodef{lrp}[LRP]{Layer-wise Relevance Propagation}
\acrodef{dnn}[DNN]{deep neural network}
\acrodef{lar}[LAR]{Learning \& Analytics for Robots}
\acrodef{rps}[RPS]{Robot Programming Suite}
\acrodef{ide}[IDE]{integrated development environment}
\acrodef{hmi}[HMI]{human-machine interaction}

\usepackage[bookmarks=false]{hyperref}
    \hypersetup{colorlinks,
      linkcolor=blue,
      citecolor=blue,
      urlcolor=blue}

\begin{document}
\begin{frontmatter}



\dochead{57th CIRP Conference on Manufacturing Systems 2024 (CMS 2024)}%

\title{Human-AI Interaction in Industrial Robotics: Design and Empirical Evaluation of a User Interface for Explainable AI-Based Robot Program Optimization}

\author[a,b]{Benjamin Alt\textsuperscript{*,}} 
\author[a]{Johannes Zahn}
\author[a]{Claudius Kienle}
\author[c]{Julia Dvorak}
\author[c]{Marvin May}
\author[a]{Darko Katic}
\author[a]{Rainer Jäkel}
\author[d]{Tobias Kopp}
\author[b]{Michael Beetz}
\author[c]{Gisela Lanza}

\address[a]{ArtiMinds Robotics, Albert-Nestler-Str. 11, 76131 Karlsruhe, Germany}
\address[b]{Institute for Artificial Intelligence, University of Bremen, Am Fallturm 1, 28359 Bremen, Germany}
\address[c]{wbk Institute of Production Science, Karlsruhe Institute of Technology, Gotthard-Franz-Straße 5, 76131 Karlsruhe, Germany}
\address[d]{Institute for Learning and Innovation in Networks, Karlsruhe University of Applied Sciences, Moltkestraße 30, 76133 Karlsruhe, Germany}

\aucores{* Corresponding author. Tel.: +49 721 509998-66. {\it E-mail address:} benjamin.alt@uni-bremen.de}

\begin{abstract}
While recent advances in deep learning have demonstrated its transformative potential, its adoption for real-world manufacturing applications remains limited. We present an \ac{xui} for a state-of-the-art deep learning-based robot program optimizer which provides both naive and expert users with different user experiences depending on their skill level, as well as Explainable AI (XAI) features to facilitate the application of deep learning methods in real-world applications. To evaluate the impact of the \ac{xui} on task performance, user satisfaction and cognitive load, we present the results of a preliminary user survey and propose a study design for a large-scale follow-up study.
\end{abstract}


\end{frontmatter}



\section{Introduction}
\label{sec:introduction}

The past two years have seen breakthrough success of \ac{dl} and related \ac{ai} approaches in a variety of domains such as natural language and image processing. In industrial robotics, \ac{ai} promises to leverage large amounts of data to automatically optimize robot programs \cite{alt_robot_2021} or enable seamless human-robot collaboration \cite{evangelou_approach_2021}, saving costs, addressing shortages of qualified labor and increasing the precision or robustness of robot-based production processes \cite{alt_heuristic-free_2022,muxfeldt_automatic_2018}. Despite this potential, \ac{ai} in the manufacturing industry remains far from widespread \cite{peres_industrial_2020}. With the introduction of digital technologies in manufacturing such as \ac{ERP} and \acp{MES}, a ``skill gap'' between the IT skills required to configure, operate and maintain the novel systems and the lack of such skills among the workforce slowed the digital transformation of the manufacturing industry \cite{azmat_closing_2020}. A similar skill gap is evident for \ac{ai} technologies, with workers lacking the required skills and experience in mathematics, programming or data science to deploy, interact with or understand state-of-the-art \ac{ai} methods \cite{gurdur_broo_rethinking_2022}. \Ac{xai} has been identified as a key component of increasing industry adoption of \ac{ai}, with interpretability tools considered a ``main catalyzer'' of industrial \ac{ai} \cite{peres_industrial_2020}. To be useful in practical applications, \ac{xai} methods must be paired with \acfp{xui} to display explanations and facilitate user interaction \cite{chromik_human-xai_2021}. Several principles for the systematic development of \acp{xui} have been proposed \cite{haid_explaining_2023}. However, comparatively few real-world \ac{xui} systems have been studied from a \ac{hmi} perspective. Studies of \acp{xui} for \ac{ai} agents in strategy games \cite{dodge_how_2018} and loan applications \cite{bove_investigating_2023} are notable examples. While in the manufacturing domain, \ac{xai} methods have been proposed for a wide range of applications such as forecasting, decision making and job scheduling \cite{chen_explainable_2023}, to our knowledge, no study of \acp{xui} for robot programming has yet been undertaken.

In this paper, we present an \ac{xui} for a state-of-the-art \ac{dl}-based method for optimizing industrial robot programs and describe its underlying design principles. We present a preliminary, small-scale user study on a realistic assembly use case, which evaluates to what extent our proposed \ac{xui} enables industry practitioners to use \ac{ai} for robot program optimization. Lastly, we propose a study design for a planned large-scale user study to evaluate the potential of \ac{xui} methods in industrial manufacturing.

\begin{figure}
\includegraphics[width=\linewidth]{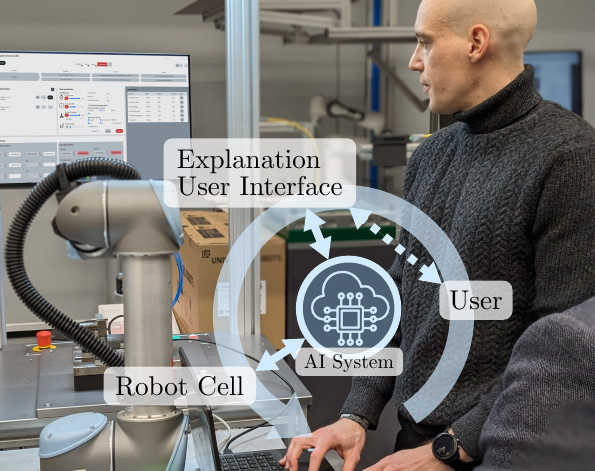}
\caption{Overview of the proposed system: A user interface enables intuitive interaction of a human user with an \ac{ai} system for robot program optimization.}
\label{fig:system_overview}
\end{figure}

\section{An Explanation User Interface for AI-based Robot Program Optimization}
Industrial robots are currently programmed by experts using textual or graphical programming interfaces. Particularly the task-specific parametrization of robot programs during the ramp-up phase of robot workcells has been identified as a major cost factor in industrial automation \cite{schmitt_future_2018}. We present an \ac{xui} for an \ac{ai}-based robot program optimizer to enable industry practitioners the use of state-of-the-art \ac{ai} methods (see Fig.~\ref{fig:system_overview}).

\subsection{AI-based Robot Program Optimization}
\label{sec:spi_workflow}
 \ac{spi} is a novel, deep-learning-based approach for optimizing robot program parameters \cite{alt_robot_2021,alt_heuristic-free_2022}. Given a parameterized robot program $P$ with parameters $x$, the execution of this program will result in a robot trajectory $\Theta$. Initially, \ac{spi} learns a \textit{shadow model} $\hat{P}$ of the program: A differentiable, neural model, which is trained on data from executions of $P$ to predict the expected execution of $P$ given some parameters $x$. It then uses $\hat{P}$ in a model-based iterative first-order optimization process to optimize the parameters $x$ to maximize a task-specific objective function. We refer to the literature for a detailed description of the approach \cite{alt_robot_2021,alt_heuristic-free_2022}. From a user's perspective, \ac{spi} requires the following three-step workflow:
\begin{enumerate}
    \item \textbf{Dataset definition:} The training data for $\hat{P}$ consists of input-label pairs $(x, \Theta)$. For \ac{ai} novices, it is challenging to determine whether a given dataset is suitable for training. Poor data quality due to e.g. outliers can severely impact optimization results. One aspect generally missed by \ac{ai} novices is the requirement for the inputs $x$ to cover the range of the parameter space over which optimization is to be performed, and to have sufficient variance. 
    \item \textbf{Model training:} $\hat{P}$ contains neural networks, which must be trained. Challenges involve the choice of appropriate hyperparameters such as learning rates or network sizes, depending on the task. After a model has finished training, it is challenging for non-experts to access the quality of the trained model.
    \item \textbf{Parameter optimization:} The program parameters are optimized with respect to a task-specific objective function. At the time of writing, \ac{spi} supports process metrics such as cycle time, path length and task success probability, as well as a threshold on the forces that are allowed to occur during program execution, and arbitrary weighted combinations of these objectives. Choosing the appropriate objective function(s) and weights is challenging for non-experts, as is the choice of appropriate hyperparameters for the optimizer. The assessment of the quality of the optimization results is likewise challenging for naive users, as the relationship between program parameters and robot behavior is not always immediately obvious, depending on the parameter. 
\end{enumerate}

\begin{figure*}
    \centering
    \includegraphics[width=\linewidth,height=.2\textheight]{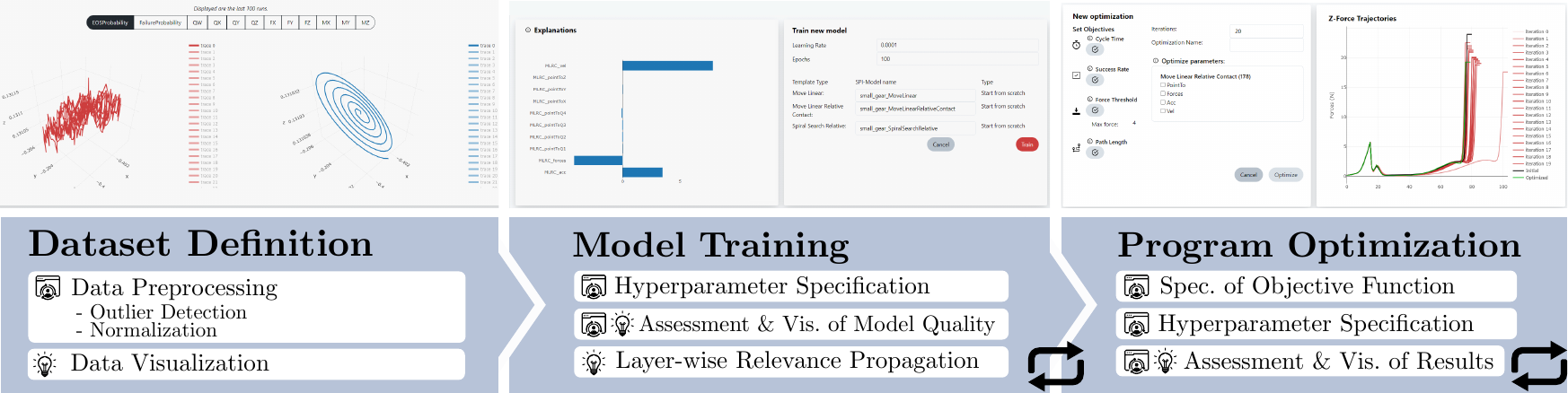}
    \caption{Workflow and corresponding \ac{ui} elements, with variation points for user adaptability (\raisebox{-.5ex}{\includegraphics[width=1em]{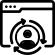}}) and explainability features (\raisebox{-.3ex}{\includegraphics[width=1em]{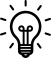}}). Screenshots for data visualization (left), \ac{lrp} (center left, bars illustrate the relative impact of individual program parameters on model output), hyperparameter specification (center right), specification of the optimization objective (2nd from right) and visualization of optimization results (far right) are shown.}
    \label{fig:ui_workflow}
\end{figure*}
 
\subsection{Guiding Principles for UI Design}
A \ac{ui} is required to enable the practical use of a complex \ac{ai} approach such as \ac{spi}. We propose that to enable \ac{ai} novices to use \ac{spi} and similar systems, \ac{ui} designers ought to emphasize \textit{user adaptability} and \textit{explainability}, in addition to general \ac{ui} design principles such as informative feedback, consistency, minimal memory load and user control \cite{ruiz_unifying_2021}.

\subsubsection{User Adaptability}
The skill gap between the level of algorithmic and data competence required to use complex \ac{ai} systems and the lack of experience of practitioners in industry with such systems often prevents the practical use of \ac{ai} \cite{gurdur_broo_rethinking_2022}. \textit{Adaptable user interfaces}, where the \ac{ui} can be reconfigured to adapt to the needs of users, can help bridge this gap \cite{schlungbaum_individual_1997}. We propose an adaptation mechanism by which the user can switch between ``Guided'' and ``Expert'' modes of the \ac{ui}. In Guided mode, several features which give the user direct control over the \ac{ai} system, such as hyperparameter selection, are simplified or hidden and replaced by defaults. \Ac{ui} elements directly visualizing technical aspects of the \ac{ai} system, such as loss curves, are simplified, removed or replaced by more intuitive, but less information-dense metrics, and additional textual explanations are added. In Expert mode, all control options, such as training and optimization hyperparameters, are exposed to the user; plots have a higher degree of information density, and all default settings can be manually overridden. Users can freely toggle between these modes.

\subsubsection{Explainability}
Explainability has been identified as a crucial factor in human-\ac{ai} interaction \cite{chromik_human-xai_2021}, significantly improving both trust in the system as well as task success \cite{leichtmann_effects_2023}. We conceptualize the proposed interface as an \ac{xui}, integrating explainability features at every step of the workflow (see Fig. \ref{fig:ui_workflow}). We take a broad view of \ac{xai} that includes both explanations of the behavior of the \acp{dnn} at the heart of the system, but also the input-output relationships of the system as a whole. The following paragraphs outline the integrated \ac{xai} features at each step of the workflow.

\subsection{XUI Workflow}
\label{sec:ui_overview}
The \ac{ui} was realized as a component of ArtiMinds \ac{lar}, a software platform for the collection and analysis of robot data. Robot programs are assumed to have been created through ArtiMinds \ac{rps}, an \ac{ide} for industrial robots based on a task-based programming paradigm. When an \ac{rps} program is executed on a robot, data such as end-effector poses, joint states or forces and torques are streamed to \ac{lar} and automatically annotated with semantic information about the currently executed robot skill, as well as user-defined tags such as serial numbers or part variants.
The \ac{ui} guides the user through the \ac{spi} workflow outlined in Section \ref{sec:spi_workflow}. Fig. \ref{fig:ui_workflow} illustrates the workflow and the realized adaptability and explanation features at each step. At the beginning of the interaction, the user selects the robot program and one or more robot skills (the \textit{target skills}) in the program to optimize (the \textit{target program}). Once a workflow step has been completed, the user can press ``Next'' to continue to the next workflow step, and can return to completed workflow steps at any time to make changes.

\begin{figure*}
    \centering
    \begin{minipage}[t]{.68\linewidth}
        \includegraphics[width=\linewidth]{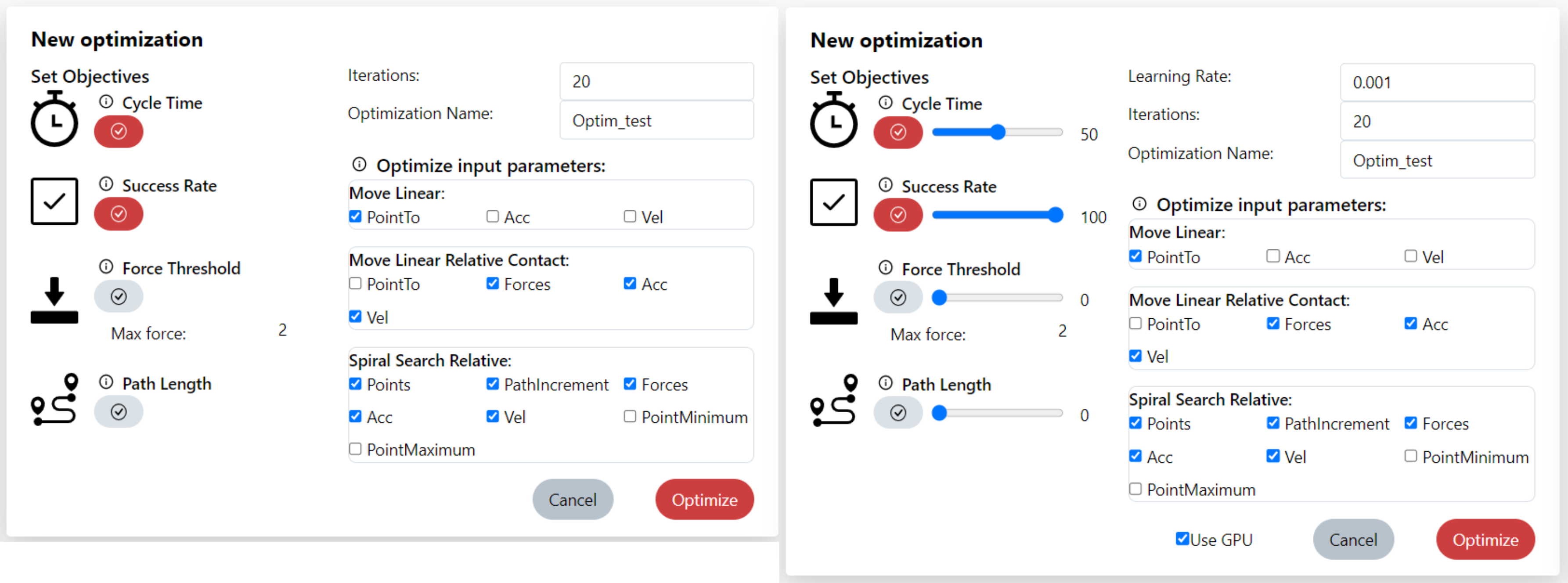}
        \caption{\ac{ui} for the optimization step, Guided (left) and Expert modes (right).}
        \label{fig:xui_optimization_guided_vs_expert}
    \end{minipage}%
    \hfill
    \begin{minipage}[t]{.31\linewidth}
        \includegraphics[width=\linewidth]{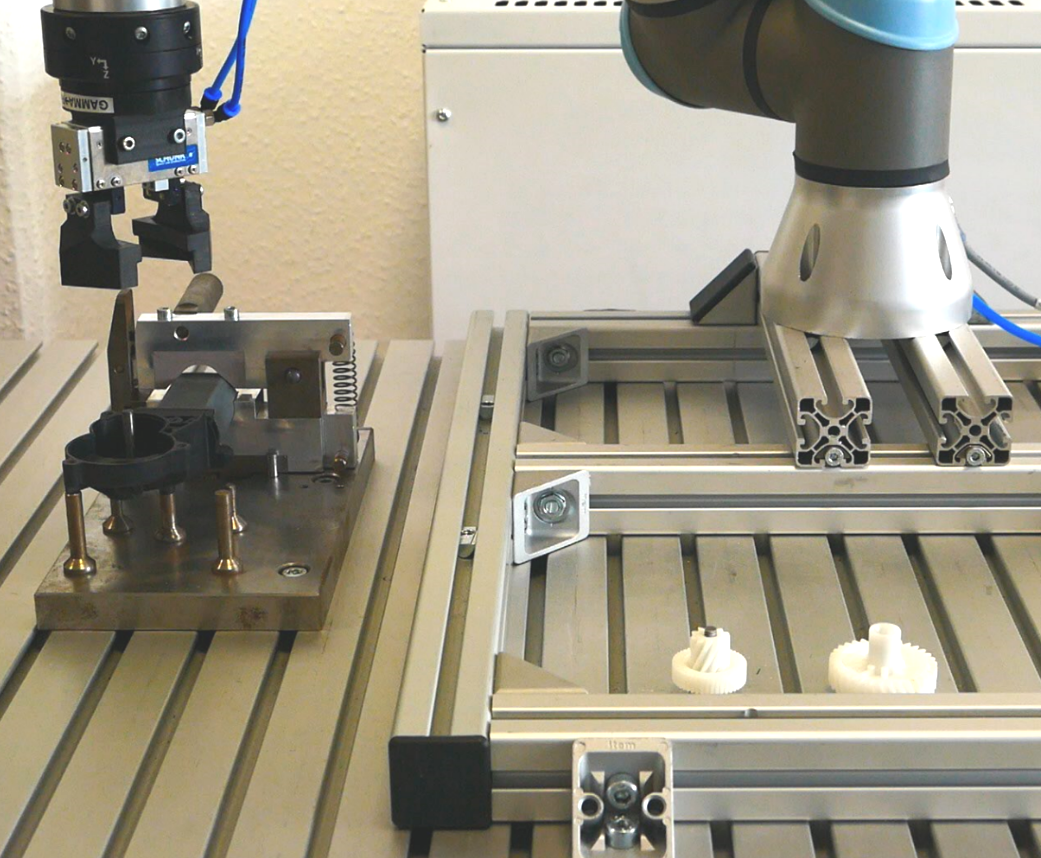}
        \caption{Experiment set-up for robotic gearbox assembly.}
        \label{fig:use_case}
    \end{minipage}
\end{figure*}

\subsubsection{Dataset definition}
To facilitate the definition of a suitable dataset for model training, the user is provided with \ac{ui} elements for exploring and visualizing collected data. To prevent the user from selecting invalid data, only data collected from executions of the target program are displayed. The user can filter training data by timestamp or any tags applied to the data at collection time. In Expert mode, the user can adjust further technical details of the dataset, such as the length with which trajectories are padded. Visualizations such as 3D plots of spatial data are displayed to provide the user with an understanding of the data. The challenges most frequently encountered when specifying datasets for \ac{spi} are a lack of variance in the inputs, the presence of outliers in the labels, as well as the definition of successful skill executions. To allow users to quickly gauge the variance in the inputs, the input distribution of the current data selection is displayed as box plots. Additionally, for each input parameter, a textual message is displayed informing the user whether the variance in that parameter is sufficient to allow later optimization of that parameter. The trajectory distribution is displayed in two plots, the first shows the end effector motion as a 3D line plot, the second displays the force-torque distribution over time. The trajectories are colored by their success, allowing easy identification of unsuccessful executions. When the user is satisfied with the data selection, they are prompted to provide a name for the dataset, which serves for later identification.

\subsubsection{Model training}
To allow the user to train shadow models for the target skills given the training data, the user is presented with a list of available pretrained models (\textit{base models}) which match the types of the target skills. For each target skill, they can choose to use a base model as-is, to finetune the base model on the previously configured dataset, or to train a new model from scratch. For each model to be trained, they are guided through a dialog to specify a name for the model for later identification, and hyperparameters for training, such as a learning rate, batch size, network size etc. Sensible defaults are provided for each setting.
After training, loss curves and training metrics are displayed, as well as graphs comparing model predictions with ground-truth labels held out from the training data.
In Guided mode, the \ac{ui} for hyperparameter specification is simplified to hide all parameters but the learning rate and training duration. Additionally the loss curves are simplified significantly, giving a summarized overview over the training process. In both modes, training results are classified into ``good performance'', ``overfitting'', ``underfitting'', ``regularization'' (dropout rate too high) and ``erroneous training data'', along with explanative descriptions.
To explain the trained network's behavior, \ac{lrp} is computed to determine the relevance of each input parameter for the model's predictions. This permits domain experts to check the model's outputs for plausibility; predicted forces, for example, should typically increase with the velocity of the motion.

\subsubsection{Program optimization}
One typical user challenge when optimizing program parameters is the specification of the objective function. Process metrics such as task success, cycle time or path length are often contradictory, and simultaneous optimization for multiple metrics may result in parameterizations which are not intuitive to users, or the optimizer may fail to converge. In Expert mode, supported task objectives can be toggled, and their relative weights specified via sliders. In Guided mode, only toggling of objectives is allowed, and weights are assigned based on sensible defaults. Expert users have additional control over the hyperparameters of the optimizer, such as the step size and number of iterations (see Figure \ref{fig:xui_optimization_guided_vs_expert}). The optimized parameterization is displayed alongside the initial parameterization, allowing a quick overview of how the program would change with optimization.
To facilitate an intuitive understanding of optimization results and the behavior of the optimizer, we provide two explanation features. The model predictions for each optimization iteration are plotted together and the best iteration is highlighted, allowing the user to gauge the plausibility of the optimization results in terms of the robot's motions. Additionally, the user can interactively change the optimized parameters and the resulting model predictions are plotted, providing an intuitive method for the user to sanity-check model predictions: If they increase the velocity of the robot, the predicted impact force should also increase and vice versa.

Model training and parameter optimization are typically performed more than once, as bad model performance such as insufficient generalization often becomes apparent once it is used for optimization. The \ac{ui} permits users to navigate back and forth between training and optimization to repeat steps or finetune hyperparameters as needed.

\section{Preliminary User Study}
To quantify the influence of a targeted system design on the cooperation between human and machine for AI-based programming, a preliminary user study is conducted. Participants are tasked to solve a practical robot programming use case using the proposed system. The use case consists of optimizing a robot program for the assembly of the gearbox of a small electric motor, in which the robot grasps a gear, approaches the axle, performs force-controlled spiral search for the exact pose of the axle and inserts the gear onto the axle (see Figure \ref{fig:use_case}). A preliminary study has been carried out and the results are discussed below (see Section \ref{sec:preliminary_study_results}). As the results are promising, a further investigation of the user interface is planned (see Section \ref{sec:representative_study}).

\subsection{Study Design}
\label{sec:preliminary_study_design}

The preliminary study comprised a total of 12 participants. While they were all familiar with robotics, 5 participants had a background in mechanical engineering, 6 in computer science, and 1 in product management. Based on their accounts of prior \ac{ai} experience, we classified 4 of the participants as \ac{ai} experts and 8 as \ac{ai} novices. All participants were males over the age of 20 with at least one university degree. Each participant was presented a short task description. No additional help was provided. In a survey subsequent to the experiment, the topics (1) success, (2) usefulness, (3) usability, (4) user adaptation, (5) cognitive load and (6) transparency were evaluated on a unipolar Likert scale ranging from 1 (``not at all'') to 5 (``completely''). Questions (1), (5.1), (5.2), (5.3) and (5.4) are a subset of the NASA TLX questionnaire \cite{hart_nasa-task_2006}. The questions on system explainability (6.1 - 6.9) are adapted from the ``\ac{xai} question bank'' \cite{liao_questioning_2020}.

\subsection{Results}
\label{sec:preliminary_study_results}

\begin{figure}
    \centering
    \includegraphics[width=\linewidth]{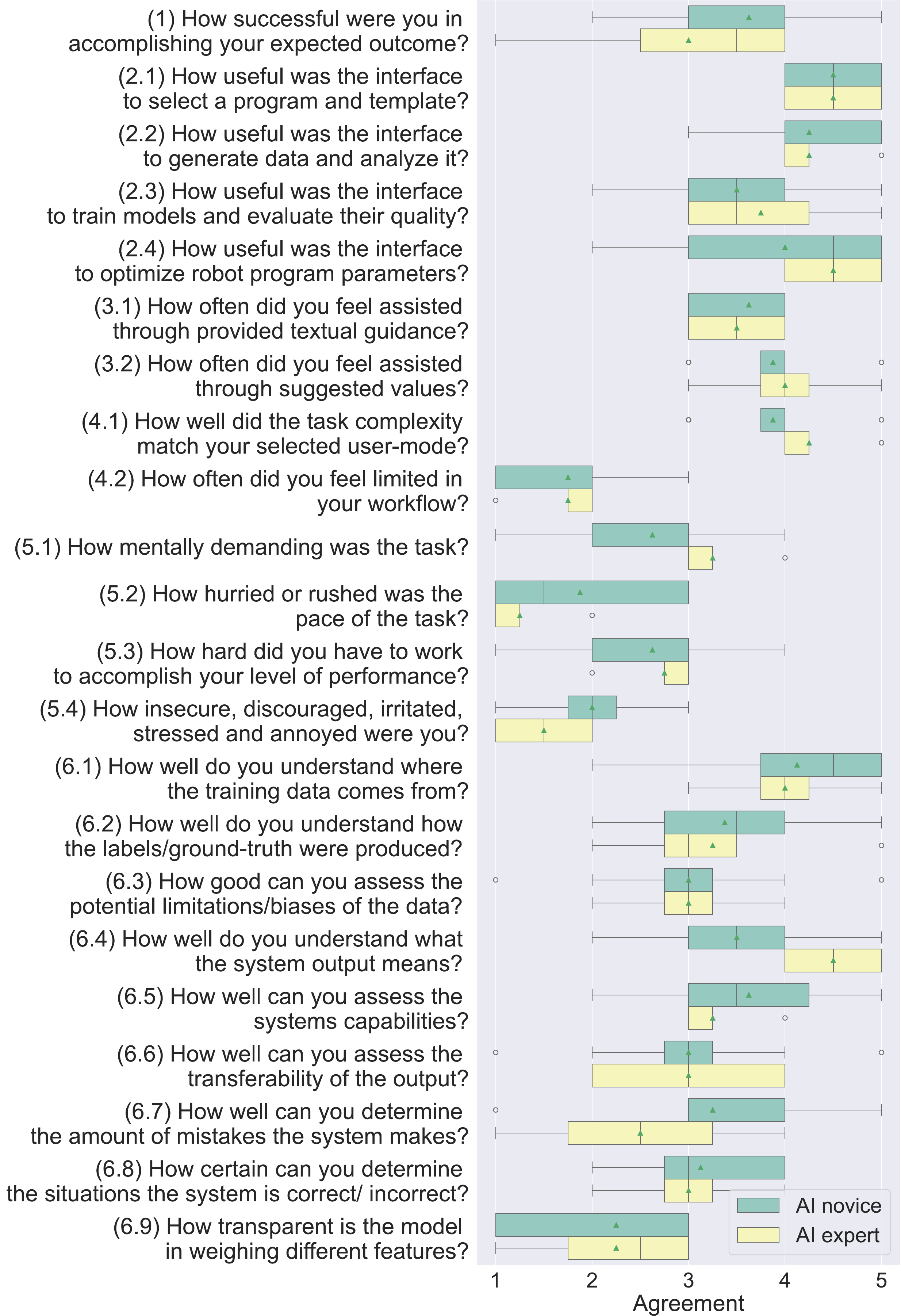}
    \caption{Survey results of 12 participants, 8 of which are classified as \ac{ai} novices and 4 as \ac{ai} experts. \textcolor[HTML]{55a868} {$\blacktriangle$} indicates the median response.}
    \label{fig:survey-results}
\end{figure}

The results of the survey are shown in Figure \ref{fig:survey-results}. First, the overall perceived success of the users is evaluated (1). It shows a high variance, which might be due to different expectations of the system's capabilities. Within the sub-tasks, perceived success is similarly high for both user groups with identical means for both the context selection task (2.1) and the data generation and analysis task (2.2). This might be due to the simplicity of the tasks, where the extra expertise of \ac{ai} experts is not required. Model training and evaluation (2.3) however requires a more comprehensive understanding of \ac{ai}, which the \ac{ai} experts leverage. This holds true for the final task of parameter optimization (2.4) as well. Participants understood each sub-task's objective and were able to generate a dataset, train a neural network and obtain predictions with that model. The provided textual explanations as well as default values were deemed useful by both participant groups (3). 

The interface left the choice of the user mode (Guided or Expert) to the users. This decision is not adjusted to the specific task, which is unknown in advance. Thus users might find themselves facing unexpectedly difficult tasks or being limited by a simplified interface. During the experiment, it could be observed that only one participant changed the mode from Expert to Guided during the optimization task, motivated by the user's curiosity. Most users decided for one mode and never changed it (4.1). This indicates that users are able to self estimate their level of experience, and that the level of complexity was adequately adapted. (4.2) targets the user's perception of limited agency. The low scores ($\mu_{\text{novice}}=\mu_{\text{expert}}=1.75$) indicate that users of both groups seldom felt limited in their workflow. \Ac{ai} novices still had enough interaction possibilities while presented a reduced version of the user interface.

The pace of work (5.1) and negative emotions like stress or insecurity (5.2) were higher for \ac{ai} novices which indicates a need for more guidance. Further, the experiment task was mentally demanding for most users of both user groups. In combination with the results on acquired guidance, it can be suggested that this might be due to the task being too complex.

To facilitate practical use of \ac{ai} in real-world applications, \ac{ai}-based decisions must be made transparent. The results showed that the source of the data was mostly understood by both user groups ($\mu_{\text{novice}}=4.125$, $\mu_{\text{expert}}=4.0$) (6.1). More complex tasks like the prediction of outputs was understood less well by \ac{ai} novices (6.4). The implemented \acp{lrp} plot was not fully understood (6.9). While the calculated feature importance score can serve as a starting point, explaining the neural network in depth remains a challenge.

\subsection{Discussion}
The preliminary study results indicate that the proposed system enables both \ac{ai} novices and experts to use an \ac{ai}-based robot program optimizer. The usefulness and intuitive usability of the system was ranked favorably, and users were able to perform a practical real-world program optimization task with acceptable cognitive load. The results showed high variance in perceived success, usefulness, and usability, which might be due to different expectations and understanding of the system's capabilities. The study emphasizes the need for additional guidance for \ac{ai} novices, as well as more advanced techniques for explaining neural network behavior. The reliability of the results from the \ac{ai} expert group is limited by the observed tendency of the expert participants to ``play'' with the system out of curiosity, rather than using the system with determination to achieve the task. Future studies will investigate methods to improve the comparability of results from different user groups.

\section{Large-Scale User Study}
\label{sec:representative_study}
In a large-scale study, the preliminary study will be systematized and the number of subjects increased to achieve statistical significance. The subjects will be students of engineering, computer science, and related disciplines, PhD candidates as well as engineers working in the field of automation or robotics.

\subsection{Study Design}
The two independent variables of the study are the level of explainability and the level of control of the system. For both, a distinction is made between low and high, resulting in a 2x2 matrix defining four variants of the user interface. We propose a double-blind between-subject design: To every participant, only one variant of the interface is presented, and neither the participants nor the experimenter know which of the four groups they are assigned to. Participants are asked to fill out a pre-task questionnaire collecting demographic data (e.g. age, gender), self-reported level of experience with \ac{ai}, industrial automation and robotics, attitude towards robots (GAToRs \cite{koverola_general_2022}), attitude towards \ac{ai} (GAAIS \cite{schepman_general_2023}) and technology commitment \cite{neyer_entwicklung_2012}.

For the explainability variable, several features have been defined and made available in the \ac{ui}. These include statistics to explain the data, explanations of the model quality, and an evaluation of the relevance of individual elements. The level of control is expressed in the extent of possible hyperparameter tuning, both for model training and trajectory optimization.

During execution, the task completion time, success rate and \ac{ui} interaction metrics such as the number of clicks are recorded. After the task, participants complete a questionnaire, including standardized measures for cognitive load (NASA TLX \cite{hart_nasa-task_2006}), quality of explanations (Explanation Satisfaction Scale \cite{hoffman_measures_2023}) and trust in the system (Trust in Automation Questionnaire \cite{korber_theoretical_2019}). The exact design of the metrics is subject to further determination. A number of 120 subjects should be reached in order to achieve sufficient statistical power.

\section{Conclusion}
We present an \ac{xui} for an \ac{ai}-based robot program optimizer, aiming to enable industry practitioners to use state-of-the-art \ac{ai} methods for practical robot programming applications. The \ac{xui} is designed to address the skill gap between the AI competence required to use complex systems and the lack of experience of practitioners in industry, as well as to foster appropriate levels of trust in the correctness and performance of the \ac{ai} system. The interface emphasizes user adaptability and explainability as guiding principles, and proposes explainability and intuitive user interaction mechanisms for each step in the workflow. We outline a preliminary user study on a realistic assembly use case and propose a study design for a planned large-scale user study. This study will focus on the level of explainability and control afforded by system, aiming to draw generalizable conclusions about the design of user interfaces for intuitive, trustworthy human-\ac{ai} interaction in the manufacturing industry.

\section*{Acknowledgements}

This work was supported by the German Ministry of Education and Research grant 02L19C255, the DFG CRC EASE (CRC \#1320) and the EU project euROBIN (grant 101070596).



\bibliography{bibliography}
\bibliographystyle{elsarticle-num}

\clearpage\onecolumn

\end{document}